\documentclass{article}

\PassOptionsToPackage{square}{natbib}
\usepackage[final]{nips_2017}

\usepackage[utf8]{inputenc} % allow utf-8 input
\usepackage[T1]{fontenc}    % use 8-bit T1 fonts
\usepackage{hyperref}       % hyperlinks
\usepackage{url}            % simple URL typesetting
\usepackage{booktabs}       % professional-quality tables
\usepackage{amsfonts}       % blackboard math symbols
\usepackage{nicefrac}       % compact symbols for 1/2, etc.
\usepackage{microtype}      % microtypography
\usepackage{amsmath} % For math

\usepackage{graphicx}
\usepackage{float}
\graphicspath{ {images/} }
\usepackage{subcaption}
\usepackage{enumitem}
\usepackage{algorithm}
\usepackage{wrapfig}
\usepackage[noend]{algpseudocode}
\DeclareMathOperator*{\argmax}{arg\,max} % thin space, limits underneath in displays

\makeatletter
\newcommand{\manuallabel}[2]{\def\@currentlabel{#2}\label{#1}}
\makeatother

\title{Learning to Factor Policies and Action-Value Functions: Factored Action Space Representations for Deep Reinforcement learning}

\author{
  Sahil Sharma \\
  Department of Computer Science\\
  Indian Institute of Technology, Madras\\
  \And
  Aravind Suresh* \\
  Department of Electrical Engineering\\
  Indian Institute of Technology, Madras\\
\AND
  Rahul Ramesh* \\
  Department of Computer Science\\
  Indian Institute of Technology, Madras\\
\And
  Balaraman Ravindran\\
  Department of Computer Science\\
  Indian Institute of Technology, Madras\\
}

\begin{document}

\maketitle

\begin{abstract}
Deep Reinforcement Learning (DRL) methods have performed well  in an increasing numbering of high-dimensional visual decision making domains.
Among all such visual decision making problems, those with discrete action spaces often tend to have underlying compositional structure in the said action space. Such action spaces often contain actions such as  {\it go left}, {\it go up} as well as {\it go diagonally up and left} (which is a composition of the former two actions). The representations of control policies in such domains have traditionally been modeled without exploiting this inherent compositional structure in the action spaces.
We propose a new learning paradigm, Factored Action space Representations (FAR)  wherein we decompose a control policy learned using a Deep Reinforcement Learning Algorithm into independent components, analogous to decomposing a vector in terms of some orthogonal basis vectors. This architectural modification of the control policy representation allows the agent to learn about multiple actions  simultaneously, while executing only one of them.  We demonstrate that FAR yields considerable improvements on top of two  DRL algorithms in Atari $2600$: FARA3C outperforms A3C (Asynchronous Advantage Actor Critic) in $9$ out of $14$ tasks and FARAQL outperforms AQL (Asynchronous $n$-step Q-Learning) in $9$ out of $13$ tasks.
\end{abstract}

\section{Introduction}

Traditional Reinforcement Learning (RL) \citep{suttonbarto} algorithms have worked with relatively simple environments (such as grid worlds) wherein policy estimates can be constructed using tabular methods or simple linear parameterizations. The state representation in such problems often consists of hand-crafted features. Recent advances in Deep Learning (DL) \citep{dl2,dl} have enabled RL methods to scale to problems with exponentially larger state spaces and even continuous action domains. This combination of RL based cost functions and DL based compositional and hierarchical representations for the state, policies and value functions has resulted in the field of Deep Reinforcement Learning (DRL).  Such DRL methods \citep{mcts,dqn, trpo, ddpg, prior, a3c, ddqn, straw, oc, unreal} perform impressively in many challenging high-dimensional sequential decision making problem such as Atari 2600 \citep{ale}, MuJoCo \citep{mujoco}, TORCS \cite{torcs} and the board game of Go \citep{alphago}.
% The prowess of such algorithms lies in the fact that they set up and subsequently solve end-to-end optimization problems to find hierarchical and compositional state representations on top of  high-dimensional observation spaces such as streams of images.

\begin{figure}[t]
 	\centering
    \includegraphics[scale=0.5]{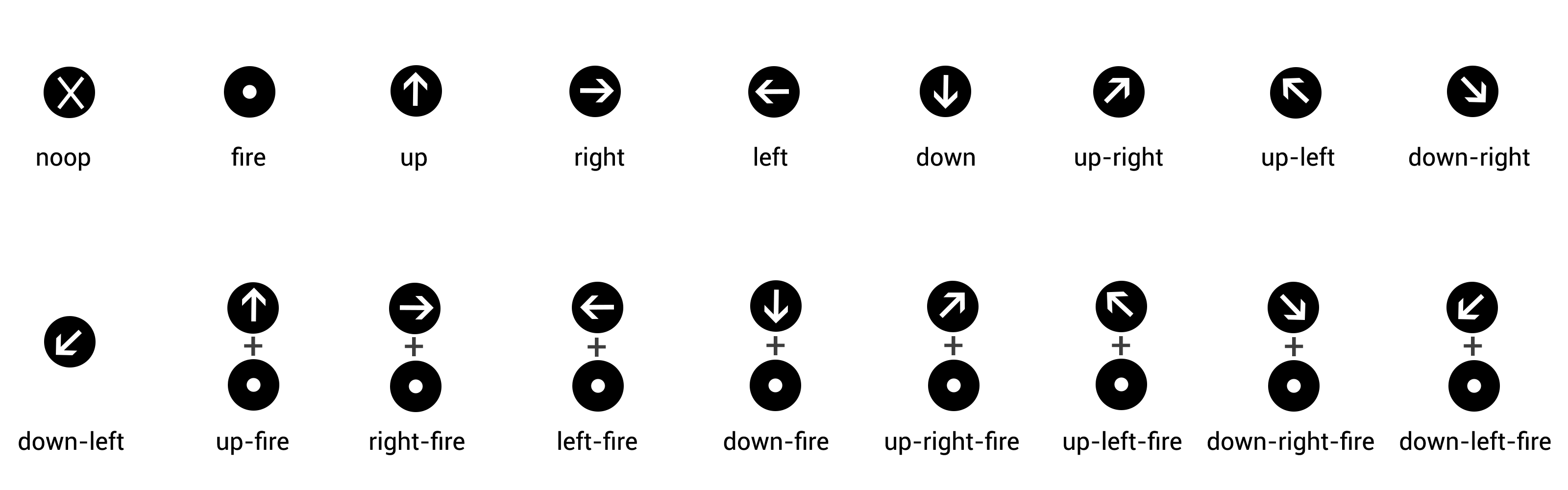}
    \caption{Visualization of the full action space for Atari $2600$}
 	\label{atari-action-set-viz}
 \end{figure}

Many such DRL algorithms operate on discrete action space domains such as Atari $2600$. The total number of actions in this domain is eighteen. These eighteen actions have been visualized in Figure \ref{atari-action-set-viz}. Observe that these eighteen actions, although presented to the DRL algorithms as the smallest indivisible units of action, are in fact not indivisible at all. There exists an inherent underlying compositional structure in this action space which a DRL algorithm can potentially take advantage of, while training. Consider a DRL agent that executes the action {\it go diagonally up and left} and gets some reward corresponding to this action. The key insight in this work is that this feedback can be used to learn not only about the {\it go diagonally up and left} action but also the actions {\it go up} and {\it go left}. Hence, every time a diagonal step is executed, it is possible to learn about the individual action factors as well. In the Atari domain, the action space can be decomposed along three independent dimensions: vertical motion (\{go up, go down, don't move vertically\}), horizontal motion (\{go left, go right, don't move horizontally\}) and firing (\{fire, don't fire\}). This work explicitly factors policies and action-value functions along these dimensions by building the factoring into the DRL agent's network architecture. The notion of factoring is not new to the DRL setup. \cite{figar} have demonstrated remarkable improvements in a variety of domains using a framework that factors the policy into one for choosing actions and another for choosing repetitions.

The key motivations for our work are principles underlying biological systems, that have shown to learn representations in an independent and orthogonal manner \citep{ortho_sound_rep,indep_edge_filters}. At an abstract level, our work is also similar to intra-option learning \citep{intra} frameworks, which revolves around the idea of learning about an option while executing another.  Arguably, humans also make decisions in a similar manner, with the fundamental unit of decisions being the orthogonal directions along which the decision space can be decomposed. While our factoring scheme can be used to extend any DRL algorithm which operates with compositional discrete action spaces, in this paper we extend the Asynchronous Advantage Actor Critic (A3C) and the Asynchronous $n$-step Q-Learning \citep{a3c} algorithms.
We demonstrate empirically that our proposed paradigm , Factored Action space Representations (FAR), provides considerable improvements over the A3C and AQL algorithms. We also provide an analysis of the factored policy learned using FARA3C that demonstrates its robustness compared to A3C policies. 

\section{Preliminaries}

\subsection{Q-learning}
Some control algorithms estimate the optimal action-value function $Q^*(s,a)$ which guides the policy executed by the RL agent. One such off-policy TD-learning algorithm \citep{td} is Q-learning \citep{watkins}. Q-learning results in the convergence of an estimated Q-function to the optimal Q-function. The Q-learning updates are give by the equation:
\[  Q(s_t,a_t) \leftarrow Q(s_t,a_{t}) + \alpha \Big( r_{t+1} + \gamma \max_{a'} Q(s_{t+1},a') - Q(s_t, a_t)\Big) \]
A policy can be derived from the Q-function estimate by selecting any action $a$ such that $a \in \argmax \limits_{a'} Q(s,a')$.

% \subsection{Advantage Actor-Critic}
% Actor-Critic algorithms \citep{ac} are policy-gradient methods \citep{suttonbarto}.  Most Actor Critic algorithms use parametric representations for the actor ($\pi(s; \theta)$) and the critic $V(s; w)$. A sample estimate of the unbiased policy gradient update is given by $ \Delta \theta_t  = \Big(\nabla _{\theta } log(\pi(a_t|s_t; \theta_t)\Big)~G_t$. Such estimates result in high variance updates to the policy parameters. $Q(s_{t}, a_{t})$ is often used as a biased estimate for $G_t$, providing the benefit of lower variance. The introduction of a state-dependent baseline $b(s_t)$ further reduces the variance  with one  popular choice for such baselines being the value-function $V(s_t; w)$. The biased low variance sample-estimate of the policy gradient is given by  $\Big(\nabla _{\theta _t} log(\pi(a_t|s_t; \theta_t)\Big)~ \Big( Q(s_t, a_t) - V(s_t) \Big)$. The second multiplicand in the sample-estimate is the known as the advantage function $A(s_t, a_t)$: 
% \[A(s_t, a_t) = Q(s_t, a_t) - V(s_t) = r_{t+1}+\gamma V(s_{t+1}) - V(s_t)\]
% where $r_{t+1}+\gamma V(s_{t+1})$ is used as an estimate for $Q(s_t,a_t)$. This estimation for $Q(s_{t}, a_{t})$ allows the agent to model the advantage function by modeling only the value function $V(s)$. 
	
\subsection{Advantage Actor-Critic}
Actor-Critic algorithms \citep{ac} are policy-gradient methods \citep{suttonbarto}.  Most Actor Critic algorithms use parametric representations for the actor ($\pi(s; \theta)$) and the critic $V(s; w)$. A biased sample-estimate for the policy gradient is given by $ \Big(\nabla _{\theta } log(\pi(a_t|s_t; \theta_t)\Big)~Q(s_{t}, a_{t})$. To lower variance in the updates, a baseline term is introduced and the sample-estimate of policy gradient becomes  $\Big(\nabla _{\theta _t} log(\pi(a_t|s_t; \theta_t)\Big)~ \Big( Q(s_t, a_t) - V(s_t) \Big)$. The second multiplicand in the sample-estimate is the known as the advantage function $A(s_t, a_t)$: 
\[A(s_t, a_t) = Q(s_t, a_t) - V(s_t) = r_{t+1}+\gamma V(s_{t+1}) - V(s_t)\]
where $r_{t+1}+\gamma V(s_{t+1})$ is used as an estimate for $Q(s_t,a_t)$. This estimation for $Q(s_{t}, a_{t})$ allows the agent to model the advantage function by modeling only the value function $V(s)$. 

\subsection{Asynchronous Advantage Actor-Critic (A3C)}
Extending the Advantage Actor-Critic algorithms naively in the DRL setup fails to work because the stochastic gradient descent algorithms commonly used with such DRL setups assume that the input samples are independent and are identically distributed. An on-policy Actor-Critic algorithm parametrized by neural networks has highly correlated inputs and therefore performs poorly when gradient based methods are used. Asynchronous Advantage Actor-Critic (A3C) methods \citep{a3c} overcome this problem by using asynchronous parallel actor-learners which simultaneously explore different parts of state space. Each learner maintains its own set of parameters which are routinely synchronized with the other learners. Using parallel actor learners ensures that their exploration of different parts of the state space translates into updates for the neural network that are relatively uncorrelated.

\subsection{Asynchronous N-step Q-learning (AQL)}
A modified version of $Q$-learning, uses an $n$-step return based TD-target \citep{peng,watkins} in order to achieve faster reward propagation and trade-off between bias and variance in the estimation of the action-value function. The modified update equation is given by:
\[ Q(s_t,a_t) \leftarrow Q(s_t,a_t) + \alpha \Big( \sum_{i=1}^n R_{t+i} + \gamma \max_{a'} Q(s_{t+n},a') - Q(s_t, a_t)\Big) \]
Similar to the Advantage Actor-Critic,  $n$-Step Q-learning, has been extended to work in the DRL setup by coming up with asynchronous $n$-step Q learning \citep{a3c}.

\section{Factored Action Representations for Deep Reinforcement Learning}
% Factored action representations ( FAR ) algorithm decomposes the actions space into independent components. The updates for the FAR structure result in changes to each of the individual factors. As a result of the same, executing an action results in updates to a set of actions that share at at least one factor. To illustrate this point, consider the 18 action-space in the Atari 2600 domain. When an action is executed, say \textit{up-right-fire}, the parameters for individual factors corresponding to up, right and fire are updated. Hence actions such  as \textit{up-fire} and \textit{right-fire} are also adjusted accordingly in the policy based on the received reward. This intuition serves as an impetus to use the FAR model.
The Factored Action Representations (FAR) framework can extend any DRL algorithm that operates on problems with compositional discrete action spaces. This includes algorithms which model $Q$-functions (like DQN \citep{dqn} or asynchronous $n$-step $Q$-learning \citep{a3c}) and actor-critic algorithms (like A3C \citep{a3c}). In the actor critic methods, FAR extends only the policy modeled by the actor; the critic is not modified. In the deep Q-learning methods, FAR modifies the representation for the action-value function that the DRL algorithm models. Let $X$ represent either the policy of an actor-critic DRL algorithm or the Q-function of a deep Q-learning algorithm.
FAR represents $X$ ($X_{i}$ corresponds to the action $a_i$) using a factored representation. Each of the factors represents a different dimension of the composite action space. We explain FAR with the Atari domain action space  as an example. Figure \ref{viz} visualizes a factoring of X over the complete Atari action space. Any action in the complete Atari action space can be represented as a tuple $(h_{i}, v_{i}, f_{i})$ with $v_{i} \in \{$go up, go down, don't move vertically$\}$, $h_{i} \in \{$go left, go right, don't move horizontally$\}$ and $f_{i} \in \{$fire, don't fire$\}$. The choice of the factors over which $X$ is decomposed depends on the set of possible actions, for a given task. This decomposed representation of $X$ allows the DRL agent to learn $X$ corresponding to multiple actions  simultaneously, while executing a single action. When the action $a =~$\textit{up-right-fire} is executed, the parameters corresponding to the individual factors of $X_{a}$: \textit{up}, \textit{right} and \textit{fire} get updated. Hence $X_{\text{up-fire}}$, $X_{\text{right-fire}}$  and $X_{\text{up-left}}$ are also adjusted, and not just the $X_{a}$.
% While executing a task, if the optimal action is $a = ~${\it fire, go left and go up}, then the actions $b = ~${\it fire and go left}, $c = ~${\it fire and go up} and $d = ~${\it go up and left} should have a high $Q$-value or probability of execution as well. In the FAR paradigm, the policies and $Q$-functions learned are constrained by this decomposition regime and have the flexibility to model this intuition.

Let $\mathcal{S}$ denote set of all states in an MDP and $\mathcal{A}$ denote the discrete action set for a DRL agent. We claim that often, action spaces $\mathcal{A}$ are compositional and thus allow the  decomposition of any action $a \in \mathcal{A}$ into $n$ independent action-factors $[a_1, a_2,\cdots, a_n]$ such that $a_i \in \mathcal{A}_i$, where $\mathcal{A}_i$ is the set of values that factor $i$ can take. We claim that instead of modeling $X$ over $\mathcal{A}$, the agent would be better off, modeling the individual components of $X$ over the factor-spaces $\mathcal{A}_{i}$. These individual components of $X$ are realized using independent output layers (referred to as factor-layers hereafter) of a neural network $f_1$, $f_2$,..., $f_n$ where $f_i$ corresponds to  $\mathcal{A}_i$  and has size $|\mathcal{A}_{i}|$. $X$ can be written in terms of the factor-layer outputs as: $X(s, a) = M(f_1(a_1|s), f_2(a_2|s),...,f_n(a_n|s))$ where $s \in \mathcal{S}$. In this equation, the combination function $M$ can have any suitable parameterized or non-parametrized form. If $X$ is a policy, then $\sum_{a \in \mathbb{A}} X(a|s) = 1$ must be enforced. Detailed training algorithms for training FAR variants proposed in this paper can be found in Appendix $B$.

\subsection{Visualization of Action Factoring for Atari}
\begin{figure}[h]
 	\centering
    \includegraphics[scale=0.4	]{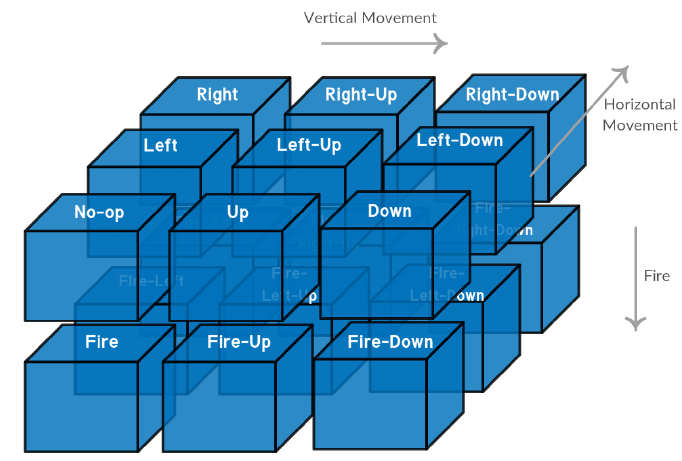}
    \caption{Visualization of action factors for Atari 2600 Domain}
 	\label{viz}
 \end{figure}

The full action space for Atari has $18$ basis actions. These basis actions can be decomposed into three independent action factors. The first factor encodes horizontal movement ($a_h = $  left/right/no horizontal movement). The second factor encodes vertical movement ($a_v = $ up/down/no vertical movement). The third factor encodes  whether to fire or not ($a_f = $ fire/don't fire). A visualization of this action space decomposition for the Atari domain is shown in Figure \ref{viz}.  This composite action space can be visualized as a 3-dimensional cuboid of dimensions $3 \times 3 \times 2$ where each $1 \times 1 \times 1$ cell represents a composite action. The axes of the cuboid represent the action-factors.
 \subsection{FARA3C}
We describe the instantiation of FAR for the A3C algorithm in this subsection. $X$ is the policy of the actor part of A3C. Let $p_h$, $p_v$ and $p_f$ denote the action-factors corresponding to horizontal, vertical and the firing dimensions. The action-factors are combined using  a {\it combination function} $M = \text{softmax} \circ m$. The action policy is given by:
\[ \pi(a|s) = \text{softmax}(m(p _h(a_h|s), p _v(a_v|s), p _f(a_f|s))) \]
In the equation,  $a = [a_h, a_v, a_f] \in \mathcal{A}$, $a_h \in \mathcal{A}_h$, $a_v \in \mathcal{A}_v$ and $a_f \in \mathcal{A}_f$. 
Figure \ref{net_vis} contains a visual depiction of FARA3C's architecture.  

\subsection{The Additive Combination Function}
The additive function is one non-parametric choice for $m$ in the definition of $M$. For FARA3C, the additive combination function has the form:
\[ \pi(a|s) = \text{softmax}(p _h(a_h|s) + p _v(a_v|s) +  p _f(a_f|s))) \]
This policy can now be re-written as a product of three probability distributions:
\[ \pi(a|s) = \text{softmax} (p _h(a_h|s)) \times \text{softmax}(p _v(a_v|s)) \times \text{softmax}(p _f(a_f|s)) \]
Consider an action $a = [a_h, a_v, a_f] \in \mathcal{A}$, such that $a_h \in \mathcal{A}_h$, $a_v \in \mathcal{A}_v$ and $a_f \in \mathcal{A}_f$. Sampling a composite action $a$ from  $\pi(a|s)$  is equivalent to independently sampling $a_{i}$ from $\pi_{i} = \text{softmax}(p_{i}(a_{i}|s))$ where  $i 
\in \{h, v, f\}$. This sub-section demonstrates that the choice of additive combination function for A3C gives the action policy a nice alternate interpretation in terms  a product of constituent factors' policies (such representations have been explored in other works such as \cite{figar}). 
% We will primarily deal with the Additive Combination function in the experiments conducted, with justification provided in section 4.1

\subsection{FARAQL}
We describe the instantiation of our framework for the AQL algorithm in this subsection. 
$X$ is the Q-function modeled by AQL. Let $Q_h$, $Q_v$ and $Q_f$ denote the set of action-factors corresponding to horizontal, vertical and the firing dimension. The action-factors are combined using  a {\it combination function} $M$. In this case:
\[ Q(s, a) = M(Q _h(a_h|s), Q _v(a_v|s), Q _f(a_f|s)) \]
where,  $a = [a_h, a_v, a_w] \in \mathcal{A}$, $a_h \in \mathcal{A}_h$, $a_v \in \mathcal{A}_v$ and $a_f \in \mathcal{A}_f$.

% NETWORK ARCHITECTURE

\begin{figure}[t]
 	\centering
 	\hspace{-10pt}
    \includegraphics[scale=0.15]{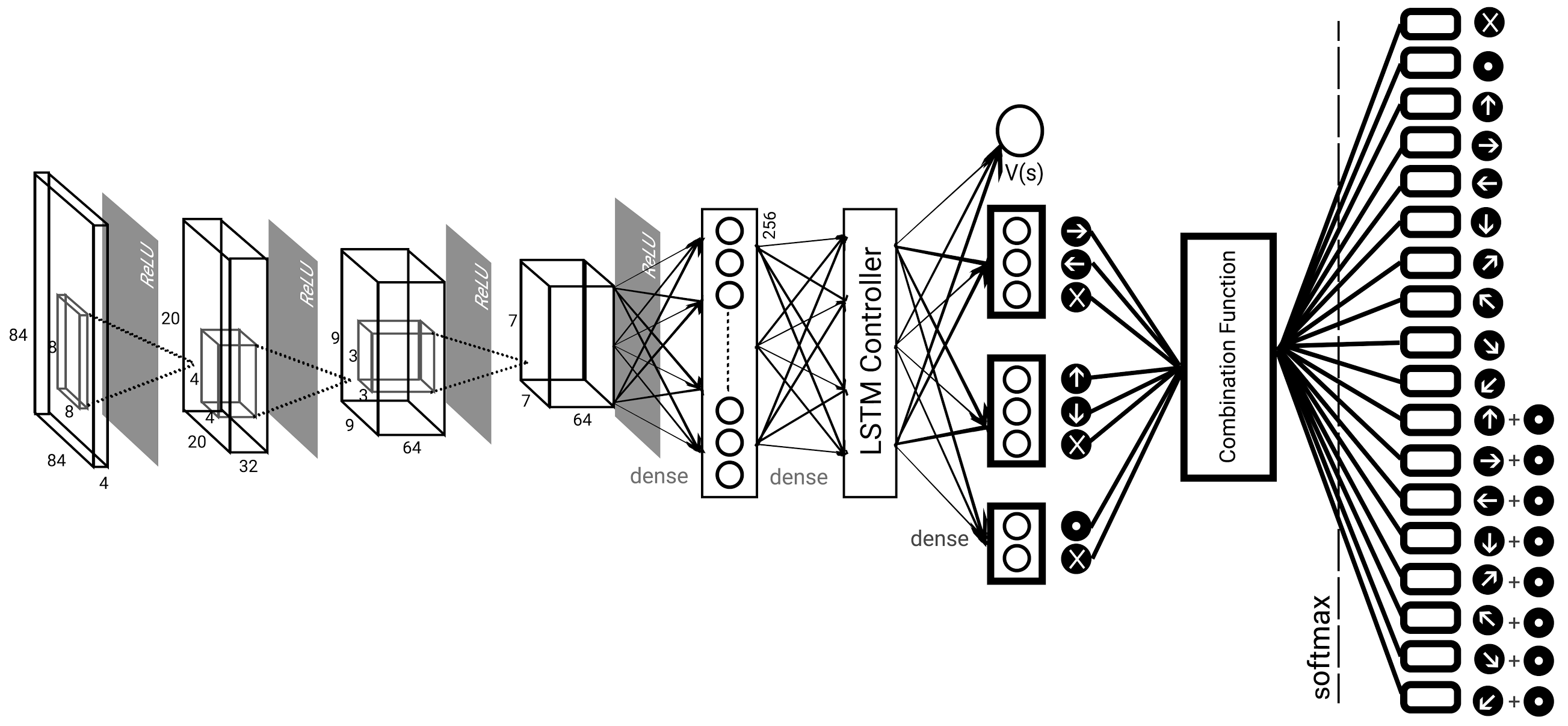}
    \caption{FARA3C's network architecture with Factored Action Representations}
	\label{net_vis}
\end{figure}

\section{Experiments and results}
\begin{figure}[ht]
 	\centering
 	\includegraphics[scale=0.52]{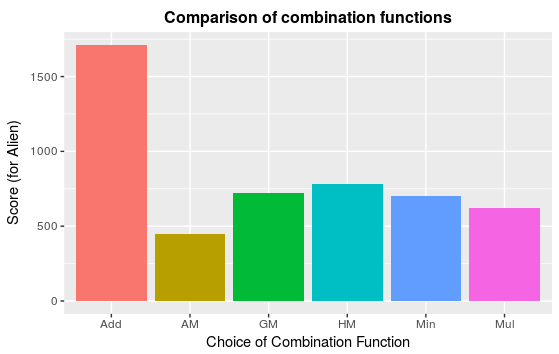}
    \caption{Comparison of the different non-parametric choices for $M$}
	\label{factor}
\end{figure}
Our experiments are geared towards answering the following questions:
\begin{enumerate}[topsep=0pt,itemsep=-1ex,partopsep=1ex,parsep=1ex]
\item{What kind of combination functions (M) work well for action space factoring?}
\item{Does action factoring improve the performance of DRL algorithms (A3C \& AQL)?}
\item{Does FAR learn fundamentally robust policy representations, compared to the baselines?}
\end{enumerate}

\subsection{Choice of the Combination Function (M)}
In this subsection we answer the first question. For chosing a good $M$, many competing FARA3C networks were trained, each implementing a different non-parametric choice of M on the \textit{Alien} task in the Atari domain. The networks were trained for $50$ million steps starting from the same random initialization. The combination functions $M$ that we experimented with were of the form $\text{softmax}(m(f_1(a_1|s), f_2(a_2|s),...,f_n(a_n|s)))$, where 
$m \in  \{\text{Summation}, \text{Multiplication}, \text{Arithmetic Mean}, \text{Harmonic Mean}, \text{Geometric Mean}, \text{Minimum}\}$. Note that the Arithmetic Mean and the Summation functions are identical barring a scaling factor of $n$ (number of action factors). This subtle difference could however lead to different policies being learned. 
% The combination function corresponding to the choice of \textit{Arithmetic Mean} can be expressed as :
% \[\pi_{am}(a|s) =  \text{softmax}\left(\left(p_v(a_v|s) + p_h(a_h|s), p_f(a_f|s\right)\right)/n)\]
% Other choices can be expressed in a similar fashion. 
\begin{figure}[t]
	\centering
    \hspace*{-0.2cm}
 	\includegraphics[scale=0.9]{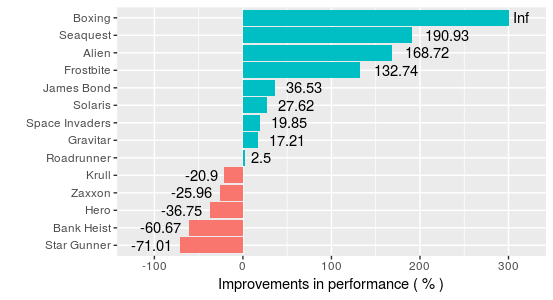}
    \caption{Comparison of the performances of FARA3C and A3C}
    \label{a3c_perf}
 \end{figure}
\begin{figure}[b]
	\centering
    \hspace{-20pt}
     \includegraphics[scale=0.7]{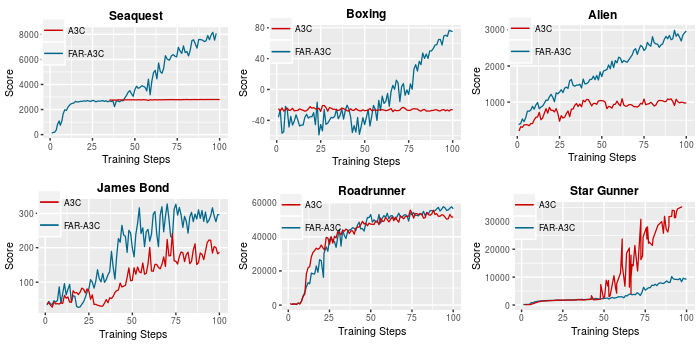}
	\caption{Training curves for FARA3C and A3C}
	 \label{a3c_training_curve_main_paper}
\end{figure}

We observed that choosing $m = \text{Summation}$  resulted in the highest performance for FARA3C as indicated by Figure \ref{factor}. We decided to stick to a similar functional form for $M$ for the AQL case as well, to demonstrate the robustness of our choice of the combination function. Hence the $Q$-values are computed using the summation of the factor-layer outputs as:
\begin{equation} 
\label{eq1}
Q(s, a) = Q _h(a_h|s) + Q _v(a_v|s) + Q _f(a_f|s))
\end{equation}

\subsection{Gameplay Experiments and Results: FARA3C}
We trained FARA3C networks as depicted in Figure \ref{net_vis} on $14$ tasks in the  Atari domain. For each of the tasks, we trained a network with three different random seeds and averaged the results to estimate the performance of the algorithm. A baseline A3C agent was also trained with the same random seeds. A comparison of FARA3C and A3C agents' performance is in Figure \ref{a3c_perf}. All the results of our experiments are tabulated in Table \ref{tab:full_a3c} in Appendix $C$. The evolution of the game-play performance on each of the tasks was plotted versus the training time.  The training curves have been averaged over the three random seeds. Figure \ref{a3c_training_curve_main_paper} contains the training curves for six of the games. Training curves for all the games can be found in Appendix $D$. Detailed explanations about the training procedure, the evaluation procedure and the hyper-parameters can be found in Appendix $A$.   
\begin{figure}[t]
	\centering
    \includegraphics[scale=0.7]{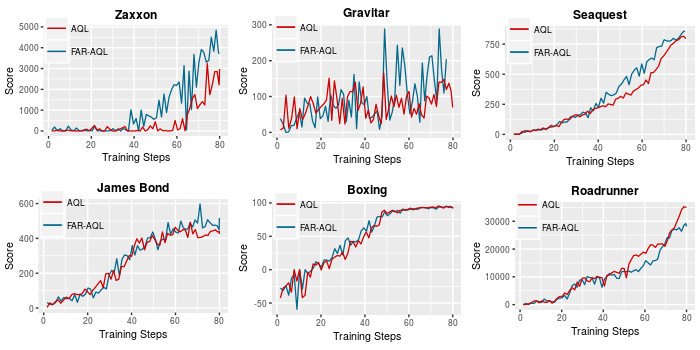}
	\caption{Training curves for AQL and FARAQL}
    \label{train_aql}
 \end{figure}

\subsection{Gameplay Experiments and Results: FARAQL}

\begin{figure}[b]
	\centering
    \hspace*{-0.2cm}
 	\includegraphics[scale=0.7]{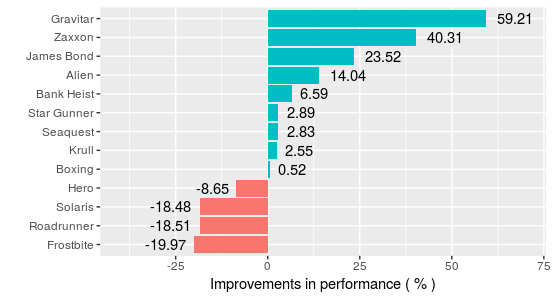}
    \caption{Comparison of the performances of FARAQL and AQL}
    \label{aql_perf}
 \end{figure}

Similar to Figure \ref{net_vis}, a modified network was constructed for FARAQL using equation \ref{eq1} and trained   for thirteen tasks in Atari domain. Performance was estimated by averaging the performances  across three random seeds. A baseline AQL agent was trained using the same random seeds. 
A visual comparison of the AQL and the FARAQL agent is presented in Figure \ref{aql_perf}. Training curves for six of the games have been presented in Figure \ref{train_aql}. Training curves for all the games can be found in Appendix $D$.
% More experimental details about the training and evaluation procedure are in Appendix B.

\begin{figure}[b]
	\centering
   \includegraphics[scale=0.57]{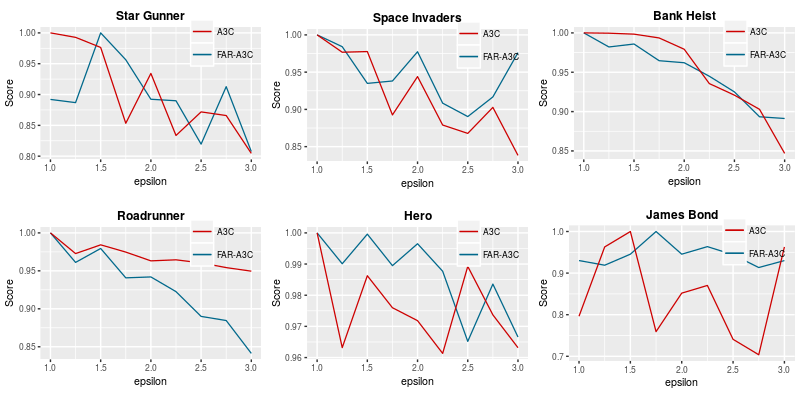}
   \caption{Variation in relative  score plotted against temperature $Z$ for the FAR and FARA3C agents}
   \label{a3c_analysis_plots_2}
\end{figure}

\section{Analysis: Test for robustness of policies learned by FARA3C \& A3C}
We claim that policies learned by FARA3C are more robust compared to those learned by A3C. This is because using a factored policy representation enables the FARA3C agents to learn about multiple actions while executing one. As a result, one would expect  FARA3C to have a better {\it ordering of actions} in the policy than A3C. In order to validate this hypothesis empirically, we conducted experiments comparing the robustness of the A3C and FARA3C policies. \subsection{Uniformly Random from best-$K$ analysis} 
\begin{figure}[t]
	\centering
   \includegraphics[scale=0.62]{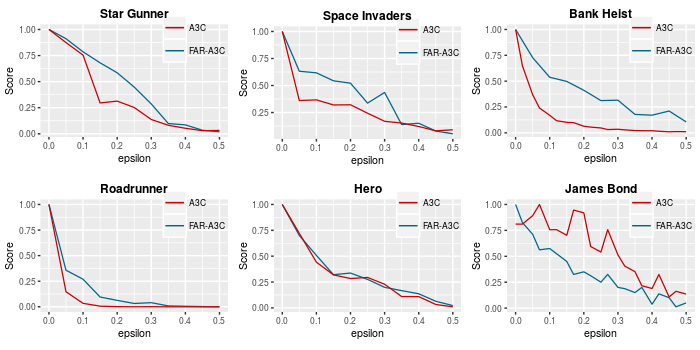}
   \caption{Variation in relative  score plotted against $\epsilon$ for the FAR and FARA3C agents}
   \label{a3c_analysis_plots}
\end{figure}

The experiment is as follows: A trained agent is taken. With probability $1 - \epsilon$, the agent samples actions from the learned policy. With probability $\epsilon$, it samples actions uniformly at random from the best $k$ actions. The use of $\epsilon$ introduces noise in the learned policy and the agents with the more robust policies would demonstrate a lower drop in performance, as $\epsilon$ increases. The value for $\epsilon$ was varied from $0$ to $0.5$ in steps of $0.05$ and the corresponding normalized performances were plotted. The individual curves have been normalized using the maximum score  to ensure a fair comparison between the FARA3C and the A3C agents by comparing only the relative changes. The experiments in this sub-section were conducted with $k=2$. From Figure \ref{a3c_analysis_plots}, observe that the FARA3C agent is comparatively more robust against policy corruption based on $\epsilon$. 
% This indicates that the $k$ actions with the largest probabilities are better for FARA3C than A3C. 

\subsection{Noise Injection analysis} 
In this analysis, a trained agent for a particular task is taken. The policy of this agent is then {\it corrupted} by injecting noise based on using the temperature ($Z$). If the policy of the agent is written as $\pi(a|s) = \text{softmax}(G(s))$, the corrupted policy $\pi_{\text{cor}}(a|s)$ of the agent can be represented as  $\pi_{\text{cor}}(a|s) = \text{softmax}( G(s)/Z)$. Increasing $Z$ results in the injection of more {\it noise} into the policy. Hence a more robust policy is expected to have a smaller drop in performance on increasing $Z$. The hyper-parameter $Z$ was varied from $1.0$ to $3.0$ in steps of $0.25$ and the performance was plotted after normalizing it against the maximum score obtained by the agent. From Figure \ref{a3c_analysis_plots_2} it is clear that for most tasks, FARA3C agents are more robust to noise injection.

\section{Conclusion and Future Work}
\label{conc_and_disc}
We propose a novel framework (FAR) for the decomposition of policies and action-value functions over a discrete action space. FAR factors  a policy/value function into independent components  and models those using independent output layers of a neural network. The FAR framework allows DRL agents to exploit the underlying compositional structure in discrete action spaces found in common RL domains to learn better policies and action-value functions. We empirically demonstrate the superiority of FAR to the baseline methods considered.  A possible extension of this framework would be to combine action-space factoring with the concept of action repetition as discussed by \cite{figar}. Action repetition could act as yet another factor of the action space and this extension could allow one to capture a large set of macro actions.

\newpage
% We used an open-source implementation of A3C available \href{http://github.com/miyosuda/async_deep_reinforce}{here} for our experiments. The Q-learning implementations were open-sourced versions which we are available here \href{https://github.com/Kaixhin/Atari}{here}.  We thank Interdisciplinary Laboratory for Data Sciences ( \href{http://web.iitm.ac.in/ilds/}{ILDS} ) for providing access to computing resources which we used for running our experiments.

\bibliography{main}

\begin{thebibliography}{26}
\providecommand{\natexlab}[1]{#1}
\providecommand{\url}[1]{\texttt{#1}}
\expandafter\ifx\csname urlstyle\endcsname\relax
  \providecommand{\doi}[1]{doi: #1}\else
  \providecommand{\doi}{doi: \begingroup \urlstyle{rm}\Url}\fi

\bibitem[Bacon et~al.(2017)Bacon, Harb, and Precup]{oc}
Bacon, Pierre-Luc, Harb, Jean, and Precup, Doina.
\newblock The option-critic architecture.
\newblock \emph{AAAI}, 2017.

\bibitem[Baumann et~al.(2011)Baumann, Griffiths, Sun, Petkov, Thiele, and
  Rees]{ortho_sound_rep}
Baumann, Simon, Griffiths, Timothy~D, Sun, Li, Petkov, Christopher~I, Thiele,
  Alexander, and Rees, Adrian.
\newblock Orthogonal representation of sound dimensions in the primate
  midbrain.
\newblock \emph{Nature neuroscience}, 14\penalty0 (4):\penalty0 423--425, 2011.

\bibitem[Bell \& Sejnowski(1997)Bell and Sejnowski]{indep_edge_filters}
Bell, Anthony~J and Sejnowski, Terrence~J.
\newblock The “independent components” of natural scenes are edge filters.
\newblock \emph{Vision research}, 37\penalty0 (23):\penalty0 3327--3338, 1997.

\bibitem[Bellemare et~al.(2013)Bellemare, Naddaf, Veness, and Bowling]{ale}
Bellemare, Marc~G., Naddaf, Yavar, Veness, Joel, and Bowling, Michael.
\newblock The arcade learning environment: An evaluation platform for general
  agents.
\newblock \emph{Journal of Artificial Intelligence Research}, pp.\  253--279,
  June 2013.

\bibitem[Bengio et~al.(2009)]{dl2}
Bengio, Yoshua et~al.
\newblock Learning deep architectures for ai.
\newblock \emph{Foundations and trends{\textregistered} in Machine Learning},
  2\penalty0 (1):\penalty0 1--127, 2009.

\bibitem[Guo et~al.(2014)Guo, Singh, Lee, Lewis, and Wang]{mcts}
Guo, Xiaoxiao, Singh, Satinder, Lee, Honglak, Lewis, Richard~L, and Wang,
  Xiaoshi.
\newblock Deep learning for real-time atari game play using offline monte-carlo
  tree search planning.
\newblock In \emph{Advances in neural information processing systems}, pp.\
  3338--3346, 2014.

\bibitem[Jaderberg et~al.(2017)Jaderberg, Mnih, Czarnecki, Schaul, Leibo,
  Silver, and Kavukcuoglu]{unreal}
Jaderberg, Max, Mnih, Volodymyr, Czarnecki, Wojciech~Marian, Schaul, Tom,
  Leibo, Joel~Z, Silver, David, and Kavukcuoglu, Koray.
\newblock Reinforcement learning with unsupervised auxiliary tasks.
\newblock \emph{To appear in 5th International Conference on Learning
  Representations}, 2017.

\bibitem[Konda \& Tsitsiklis(2000)Konda and Tsitsiklis]{ac}
Konda, Vijay~R and Tsitsiklis, John~N.
\newblock Actor-critic algorithms.
\newblock In \emph{Advances in neural information processing systems}, pp.\
  1008--1014, 2000.

\bibitem[LeCun et~al.(2015)LeCun, Bengio, and Hinton]{dl}
LeCun, Yann, Bengio, Yoshua, and Hinton, Geoffrey.
\newblock Deep learning.
\newblock \emph{Nature}, 521\penalty0 (7553):\penalty0 436--444, 2015.

\bibitem[Lillicrap et~al.(2015)Lillicrap, Hunt, Pritzel, Heess, Erez, Tassa,
  Silver, and Wierstra]{ddpg}
Lillicrap, Timothy~P, Hunt, Jonathan~J, Pritzel, Alexander, Heess, Nicolas,
  Erez, Tom, Tassa, Yuval, Silver, David, and Wierstra, Daan.
\newblock Continuous control with deep reinforcement learning.
\newblock \emph{arXiv preprint arXiv:1509.02971}, 2015.

\bibitem[Mnih et~al.(2013)Mnih, Kavukcuoglu, Silver, Graves, Antonoglou,
  Wierstra, and Riedmiller]{dqnnips}
Mnih, Volodymyr, Kavukcuoglu, Koray, Silver, David, Graves, Alex, Antonoglou,
  Ioannis, Wierstra, Daan, and Riedmiller, Martin.
\newblock Playing atari with deep reinforcement learning.
\newblock \emph{arXiv preprint arXiv:1312.5602}, 2013.

\bibitem[Mnih et~al.(2015)Mnih, Kavukcuoglu, Silver, Rusu, Veness, Bellemare,
  Graves, Riedmiller, Fidjeland, Ostrovski, Petersen, Beattie, Sadik,
  Antonoglou, King, Kumaran, Wierstra, Legg, and Hassabis]{dqn}
Mnih, Volodymyr, Kavukcuoglu, Koray, Silver, David, Rusu, Andrei~A., Veness,
  Joel, Bellemare, Marc~G., Graves, Alex, Riedmiller, Martin, Fidjeland,
  Andreas~K., Ostrovski, Georg, Petersen, Stig, Beattie, Charles, Sadik, Amir,
  Antonoglou, Ioannis, King, Helen, Kumaran, Dharshan, Wierstra, Daan, Legg,
  Shane, and Hassabis, Demis.
\newblock Human-level control through deep reinforcement learning.
\newblock \emph{Nature}, February 2015.

\bibitem[Mnih et~al.(2016{\natexlab{a}})Mnih, Agapiou, Osindero, Graves,
  Vinyals, Kavukcuoglu, et~al.]{straw}
Mnih, Volodymyr, Agapiou, John, Osindero, Simon, Graves, Alex, Vinyals, Oriol,
  Kavukcuoglu, Koray, et~al.
\newblock Strategic attentive writer for learning macro-actions.
\newblock \emph{arXiv preprint arXiv:1606.04695}, 2016{\natexlab{a}}.

\bibitem[Mnih et~al.(2016{\natexlab{b}})Mnih, Badia, Mirza, Graves, Lillicrap,
  Harley, Silver, and Kavukcuoglu]{a3c}
Mnih, Volodymyr, Badia, Adria~Puigdomenech, Mirza, Mehdi, Graves, Alex,
  Lillicrap, Timothy~P, Harley, Tim, Silver, David, and Kavukcuoglu, Koray.
\newblock Asynchronous methods for deep reinforcement learning.
\newblock In \emph{International Conference on Machine Learning},
  2016{\natexlab{b}}.

\bibitem[Peng \& Williams(1996)Peng and Williams]{peng}
Peng, Jing and Williams, Ronald~J.
\newblock Incremental multi-step q-learning.
\newblock \emph{Machine learning}, 22\penalty0 (1):\penalty0 283--290, 1996.

\bibitem[Schaul et~al.(2015)Schaul, Quan, Antonoglou, and Silver]{prior}
Schaul, Tom, Quan, John, Antonoglou, Ioannis, and Silver, David.
\newblock Prioritized experience replay.
\newblock \emph{4th International Conference on Learning Representations},
  2015.

\bibitem[Schulman et~al.(2015)Schulman, Levine, Moritz, Jordan, and
  Abbeel]{trpo}
Schulman, John, Levine, Sergey, Moritz, Philipp, Jordan, Michael~I, and Abbeel,
  Pieter.
\newblock Trust region policy optimization.
\newblock \emph{CoRR, abs/1502.05477}, 2015.

\bibitem[Sharma et~al.(2017)Sharma, Lakshminarayanan, and Ravindran]{figar}
Sharma, Sahil, Lakshminarayanan, Aravind~S., and Ravindran, Balaraman.
\newblock Learning to repeat: Fine grained action repetition for deep
  reinforcement learning.
\newblock \emph{To appear in 5th International Conference on Learning
  Representations}, 2017.

\bibitem[Silver et~al.(2016)Silver, Huang, Maddison, Guez, Sifre, van~den
  Driessche, Schrittwieser, Antonoglou, Panneershelvam, Lanctot, Dieleman,
  Grewe, Nham, Kalchbrenner, Sutskever, Lillicrap, Leach, Kavukcuoglu, Graepel,
  and Hassabis]{alphago}
Silver, David, Huang, Aja, Maddison, Chris~J., Guez, Arthur, Sifre, Laurent,
  van~den Driessche, George, Schrittwieser, Julian, Antonoglou, Ioannis,
  Panneershelvam, Vedavyas, Lanctot, Marc, Dieleman, Sander, Grewe, Dominik,
  Nham, John, Kalchbrenner, Nal, Sutskever, Ilya, Lillicrap, Timothy~P., Leach,
  Madeleine, Kavukcuoglu, Koray, Graepel, Thore, and Hassabis, Demis.
\newblock Mastering the game of go with deep neural networks and tree search.
\newblock \emph{Nature}, 529\penalty0 (7587):\penalty0 484--489, 2016.
\newblock \doi{10.1038/nature16961}.
\newblock URL \url{http://dx.doi.org/10.1038/nature16961}.

\bibitem[Sutton(1988)]{td}
Sutton, Richard~S.
\newblock Learning to predict by the methods of temporal differences.
\newblock \emph{Machine learning}, 3\penalty0 (1):\penalty0 9--44, 1988.

\bibitem[Sutton \& Barto(1998)Sutton and Barto]{suttonbarto}
Sutton, Richard~S. and Barto, Andrew~G.
\newblock Introduction to reinforcement learning.
\newblock \emph{MIT Press}, 1998.

\bibitem[Sutton et~al.(1998)Sutton, Precup, and Singh]{intra}
Sutton, Richard~S, Precup, Doina, and Singh, Satinder~P.
\newblock Intra-option learning about temporally abstract actions.
\newblock In \emph{ICML}, volume~98, pp.\  556--564, 1998.

\bibitem[Todorov et~al.(2012)Todorov, Erez, and Tassa]{mujoco}
Todorov, Emanuel, Erez, Tom, and Tassa, Yuval.
\newblock Mujoco: {A} physics engine for model-based control.
\newblock In \emph{2012 {IEEE/RSJ} International Conference on Intelligent
  Robots and Systems, {IROS} 2012, Vilamoura, Algarve, Portugal, October 7-12,
  2012}, pp.\  5026--5033, 2012.
\newblock \doi{10.1109/IROS.2012.6386109}.
\newblock URL \url{http://dx.doi.org/10.1109/IROS.2012.6386109}.

\bibitem[Van~Hasselt et~al.(2016)Van~Hasselt, Guez, and Silver]{ddqn}
Van~Hasselt, Hado, Guez, Arthur, and Silver, David.
\newblock Deep reinforcement learning with double q-learning.
\newblock In \emph{AAAI}, pp.\  2094--2100, 2016.

\bibitem[Watkins \& Dayan(1992)Watkins and Dayan]{watkins}
Watkins, Christopher J. C.~H. and Dayan, Peter.
\newblock Technical note: Q-learning.
\newblock \emph{Mach. Learn.}, 8\penalty0 (3-4):\penalty0 279--292, May 1992.

\bibitem[Wymann et~al.(2000)Wymann, Espi{\'e}, Guionneau, Dimitrakakis, Coulom,
  and Sumner]{torcs}
Wymann, Bernhard, Espi{\'e}, Eric, Guionneau, Christophe, Dimitrakakis,
  Christos, Coulom, R{\'e}mi, and Sumner, Andrew.
\newblock Torcs, the open racing car simulator.
\newblock \emph{Software available at http://torcs. sourceforge. net}, 2000.

\end{thebibliography}
\bibliographystyle{nips}
\newpage

\section*{Appendix A: Experimental Details}
In this appendix, we document the experimental details for all of our experiments. Note that all the reported game-play performances and graphs are averages across $3$ random seeds to ensure that the comparisons are robust to random starting points for the parameter vectors. 
\label{appendix_a}

\subsection*{FARA3C and A3C}
The human-starts evaluation paradigm proposed by \cite{a3c} is hard to replicate in the absence of the same human-tester trajectories. Hence, we use the same training and evaluation procedure as \cite{figar}.
\subsubsection*{On hyper-parameters}
We used the LSTM-variant of A3C [\cite{a3c}] algorithm for the both the FARA3C and the A3C  experiments. The async-rmsprop algorithm [\cite{a3c}] was used for updating parameters with the same hyper-parameters as in \cite{a3c}. The initial learning rate used was $10^{-3}$ and it was linearly annealed to $0$ over 100 million steps. The $n$ used in $n$-step returns was $20$. Entropy regularization was used to encourage exploration, similar to \cite{a3c}. The $\beta$ for entropy regularization was found to be $0.01$ after hyper-parameter tuning, both for FARA3C and A3C. The $\beta$ was tuned in the set $\{0.01, 0.02\}$. The optimal learning rate was found to be $7 \times 10^{-4}$ for both FARA3C and A3C separately. The learning rate was tuned over the set $\{7\times 10^{-4}, 10^{-3}\}$. The discounting factor for rewards was retained at $0.99$ since it seems to work well for a large number of methods \citep{a3c,figar,unreal}.

All the models were trained for $100$ million time steps.  Evaluation was done after every $1$ million steps of training and followed the strategy described in \cite{figar}.  This evaluation was done for $100$ episodes, with each episode's length capped at $20000$ steps, to arrive at an average score. The evolution of this average game-play performance with training progress has been demonstrated for a few games in Figure \ref{a3c_training_curve_main_paper}. An expanded version of the figure for all the games can be found in Appendix $D$. \\
Table \ref{tab:full_a3c} in Appendix $C$ contains the raw scores obtained by FARA3C and A3C agents on $15$ Atari $2600$ games. The evaluation was done using the best agent obtained after training for $100$ million steps (where the best agent was chosen according to the model selection paradigm outlines in 
cite{dqn}). 

\subsection*{Architecture details}
We used a low level architecture similar to \cite{a3c,figar} which in turn uses the same low level architecture as \cite{dqn}. Figure \ref{net_vis} contains a visual depiction of the network used for FARA3C. The same architecture was used for FARA3C and A3C agents and has been described below:\\
The first three layers  are convolutional layers with same filter sizes, strides, padding and number of filters as \cite{dqn,a3c,figar}. These convolutional layers are followed by two fully connected (FC) layers and an LSTM layer. A policy and a value function are derived from the LSTM outputs using two different output heads. The number of neurons in each of the FC layers and the LSTM layers is $256$. 

Similar to \cite{a3c} the Actor and Critic networks share all but the final output layer. Each of the two functions: policy and value function are realized with a different final output layer, with the value function outputs having no non-linearity  and with the policy and having a softmax-non linearity as output non-linearity, to model the multinomial distribution.

\newpage

\subsection*{FARAQL and AQL}
We used a highly tuned open-source implementation of asynchronous Q-learning found at:\\ \url{https://github.com/Kaixhin/Atari}.

\subsubsection*{AQL}
The learning rate was set to be $7 \times 10^{-4}$. The value of $n$ for $n$-step returns was set to be $5$. In keeping with the $\epsilon$-greedy strategy highlighted for training asynchronous $n$-step Q-learning in \cite{a3c} we randomly sampled $\epsilon$ in each thread and the $\epsilon$ was decayed to various different values from the set $\{0.05, 0.1, 0.5\}$. The training period lasted $80$ million time steps.  The starting value of epsilon was $1$ and it was linearly decayed over $64$ million steps to its final value.  The same network architecture as  \cite{dqnnips} was used for all the experiments.
\subsubsection*{FARAQL}
To better understand the generalization capabilities of our FAR framework in general and FARQL algorithm in particular, we decided to use the exact same hyper-parameter choices as AQL for FARAQL. None of the hyper-parameters for FARAQL were tuned specifically for this algorithm. This represents a hyper-parameter setting which is not very favorable to FARAQL. Even in this setting, we observe that FARAQL significantly outperforms AQL. Tuning the hyper-parameters specifically for FARAQL could result in a further improvement in performance.  

\newpage
\section*{Appendix B: The FARA3C Algorithm}

In this appendix, we present  pseudo-code versions of the training algorithm for FARA3C and FARAQL.
\subsection*{FARA3C}
\begin{algorithm}[h]
\caption{FARA3C}
\begin{algorithmic}[1]
\State // Assume global shared parameter vectors $\theta$ and $w$
\State // Assume global step counter (shared) $T = 0$
\\
\State $n$~~$\gets$~~~Number of independent factors
\State $T_{max}$~~~$\gets$~~~Total number of training steps for the FARA3C agent
\State $\pi$~~~$\gets$~~~Policy of the agent
\State $f$~~~$\gets$~~~List of factor heads for independent factors
\State Initialize local thread's step counter $t \leftarrow 1$ \\
\Repeat
\State $t _{init} = t$
\State $d\theta \leftarrow 0$
\State $dw \leftarrow 0$
\State // Assume local thread's parameters as $\theta'$ and $w'$
\State Synchronize local thread parameters $\theta' = \theta$ and $w' = w$
\State Obtain state $s_{t_{\text{init}}}$

\Repeat
	\State $\pi(a|s_{t}) = M(f_1(a_1|s_{t}), f_2(a_2|s_{t}), \cdots, f_n(a_n|s_{t}))$

\State Sample $a _t \sim \pi(a|s_t; \theta')$
\State Execute action $a_t$ to obtain reward $r_t$ and observe next state $s_{t+1}$
\State $t \leftarrow t+1$
\State $T \leftarrow T+1$
\State Obtain state $s_t$
\Until $s_t$ is terminal or $t == t_{\text{init}} + t_{\text{max}}$
\If{$s_t$ is terminal}
	\State $R$~~$\gets$~~~$0$
\Else
	\State $R \leftarrow V(s_t, w')$
\EndIf
\For{$i \in \{ t-1, \dots t_{\text{init}} \}$} 
	\State $R \leftarrow r_i + \gamma R$ 
    \State Accumulate gradients for $\theta'$: $d\theta \leftarrow d \theta + \nabla_{\theta'} log(\pi(a_i | s_i ; \theta')(R - V(s_i;w'))$  
	\State Accumulate gradients for $w'$: $dw 			\leftarrow dw + \frac{\partial (R - 		V(s_i;w'))^2}		{\partial w'}$ 
\EndFor
\State Perform asynchronous update of $\theta$ using $d\theta$ and $w$ using $dw$
    \Until{$T > T_{max}$}
\end{algorithmic}
\end{algorithm}

\newpage

\subsection*{FARAQL}

\begin{algorithm}[h]
\caption{FARAQL}
\begin{algorithmic}[1]
\State // Assume global shared parameter vector $\theta$
\State // Assume global shared parameter vector for target network $\theta^{-}$
\State // Assume global step counter (shared) $T = 0$\\
\State $n$~~$\gets$~~~Number of independent factors
\State $T_{max}$~~~$\gets$~~~Total number of training steps for the FARAQL agent
\State $Q$~~~$\gets$~~~Action-value function of the agent
\State $q$~~~$\gets$~~~List of factor heads for independent factors\\
\State Initialize local thread's step counter $t \leftarrow 1$
\State Initialize target network parameters $\theta^- \leftarrow \theta$
\State Initialize local thread's parameter $\theta' = \theta$
\Repeat
\State $t _{init} = t$
\State $d\theta \leftarrow 0$
\State $dw \leftarrow 0$
\State Synchronize local thread parameters $\theta' = \theta$
\State Obtain state $s_{t_{init}}$
\Repeat
\State $Q(s_{t}, a) = M(Q_1(a_1|s_{t}), Q_2(a_2|s_{t}), \cdots, Q_n(a_n|s_{t}))$
    
\State $a_t \sim$ the $\epsilon-\text{greedy policy}$ derived from $Q(s_t, a; \theta')$
\State Execute action $a_t$ to obtain reward $r_t$ and observe next state $s_{t+1}$
\State $t \leftarrow t+1$
\State $T \leftarrow T+1$
\State Obtain state $s_{t}$
\Until $s_t$ is terminal or $t == t_{\text{init}} + t_{\text{max}}$
\If{$s_t$ is terminal}
	\State $R$~~$\gets$~~~$0$
\Else
	\State $R \leftarrow \max_{a} (s_t, a; \theta^-)$
\EndIf
\For{$i \in \{ t-1, \dots t_{\text{init}} \}$} 
	\State $R \leftarrow r_i + \gamma R$ 
	\State Accumulate gradients for $\theta'$: $d\theta \leftarrow d \theta + \nabla_{\theta'} ((R - Q(s_i, a_i;\theta'))^2$ 
\EndFor
\State Perform asynchronous update of $\theta$ using $d\theta$
    \Until{$T > T_{max}$}
\end{algorithmic}
\end{algorithm}

\newpage
\section*{Appendix C: Raw Performance tables}

In this appendix, we document the raw performances of all our agents and the corresponding baseline performances. All the performances have been obtained by averaging across $3$ random seeds to ensure a robust comparison. 
\subsection*{FARA3C and A3C}
Table \ref{tab:full_a3c} demonstrates that FARA3C outperforms A3C on $9$ out of $14$ tasks that we experimented with. 
 \begin{table}[h]
\begin{center}
\begin{tabular}{lll}
\multicolumn{1}{c}{\bf Name}  
&\multicolumn{1}{c}{\bf FARA3C}
&\multicolumn{1}{c}{\bf A3C}
\\ \hline \\
Alien & \textbf{3033.1} & 1128.73 \\
Bank Heist & 695.9 &{\bf 1769.43} \\
Boxing & \textbf{66.00} & -26.27 \\
% Chopper Command & 4244.67 & \textbf{5113.67} \\
% Defender & 55993 & \textbf{60718.33}\\
Frostbite & \textbf{1300.70} & 558.87 \\
Gravitar & \textbf{178.17} & 152.00 \\
Hero & 15574.37 & \textbf{24623.97} \\
James Bond & {\bf 373.17} &  273.33\\
Krull & 4550.23 &{\bf 5752.17} \\
Roadrunner & \textbf{59025.00} & 57585.33 \\
Seaquest & \textbf{8159.90} & 2804.73 \\
Solaris & \textbf{4074.4} & 3192.53 \\
Space Invaders & \textbf{2056.7833} & 1716.1666   \\
Star Gunner & 10357.33 & \textbf{35725.33}\\
Zaxxon & 952.67 &  {\bf 1286.67}\\

\end{tabular}
\end{center}
\caption{A3C/FARA3C Experiments on Atari 2600 domain}
\label{tab:full_a3c}
\end{table}

\subsection*{FARAQL and AQL}
Table \ref{tab:full_aql} demonstrates that FARAQL outperforms AQL on $9$ out of $13$ tasks that we experimented with. 
 \begin{table}[h]
\begin{center}
\begin{tabular}{lll}
\multicolumn{1}{c}{\bf Name}  
&\multicolumn{1}{c}{\bf FARAQL}
&\multicolumn{1}{c}{\bf AQL}
\\ \hline \\
Alien & \textbf{2361.69} & 2070.97 \\
Bank Heist & {\bf 878.61} & 824.28 \\
Boxing & \textbf{95.47} & 94.98 \\
Frostbite & 1656.34 & {\bf 2069.62} \\
Gravitar & \textbf{403.81} & 253.64 \\
Hero & \textbf{13404.56} & 12244.68 \\
James Bond & {\bf 646.54} &  523.45\\
Krull & {\bf 10370.19} & 10112.62 \\
Roadrunner & 29656.46 & {\bf 36391.57} \\
Seaquest & \textbf{4258.37} & 4141.18 \\
Solaris & 2233.29 & {\bf 2739.61} \\
Star Gunner &{\bf 37314.48} & 36265.23  \\
Zaxxon & {\bf 5122.07} &  3650.64\\

\end{tabular}
\end{center}
\caption{AQL/FARAQL Experiments on Atari 2600 domain}
\label{tab:full_aql}
\end{table}

\newpage

\section*{Appendix D: Training curves}
This appendix contains all the training curves for all the algorithms presented in this work. All training curves have been averaged across three random seeds to ensure a robust comparison.

\begin{figure}[h]
	\centering
   \includegraphics[scale=0.77]{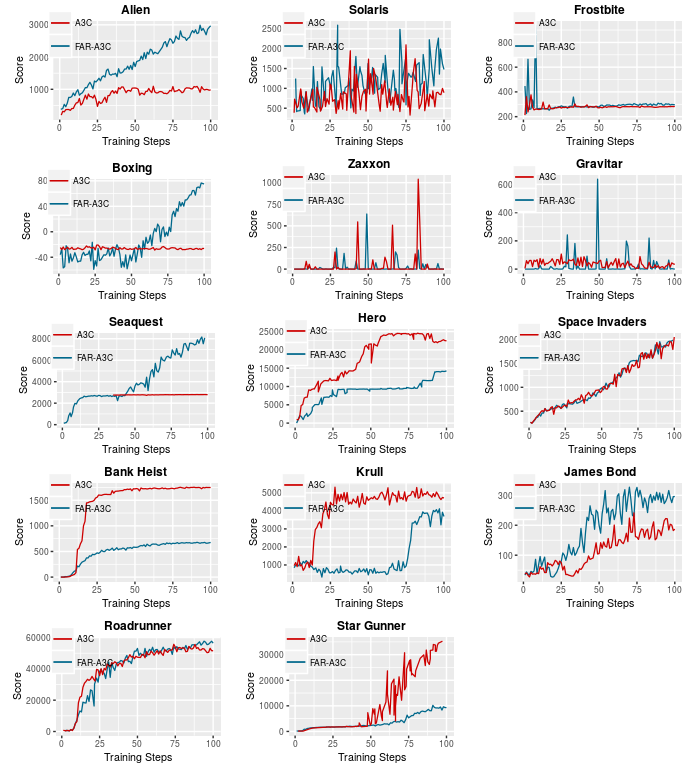}
   \caption{Training curves for the A3C and FARA3C agents}
   \label{a3c_train_plots}
\end{figure}

\newpage

\begin{figure}[h]
	\centering
   \includegraphics[scale=0.77]{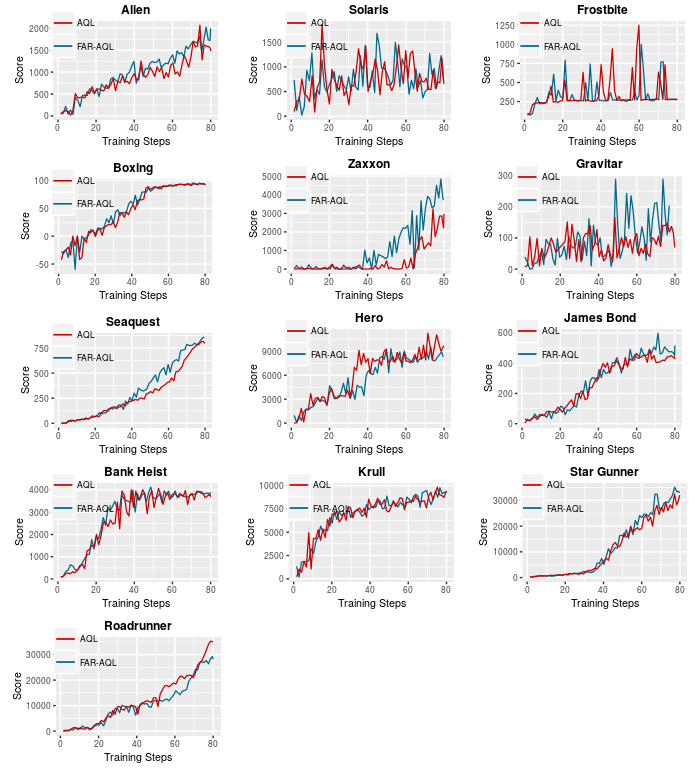}
   \caption{Training curves for the AQL and FARAQL agents}
   \label{ql_train_plots}
\end{figure}

% \newpage

% \section*{Appendix E: Additional Analysis of FARA3C agents}
% This appendix contains analysis for all the tasks on which we experimented with FARA3C. 

\end{document}